\documentclass[letterpaper, 10 pt, conference]{ieeeconf}  

\IEEEoverridecommandlockouts                              

\usepackage{url}
\usepackage{times}
\usepackage{epsfig}
\usepackage{graphicx}
\usepackage{amsmath}
\usepackage{amssymb}
\usepackage{textgreek}
\usepackage{booktabs}
\usepackage[skip=0pt]{caption} 
\usepackage{chapterbib}
\usepackage{subcaption}
\usepackage{hyperref}
\usepackage{float}

\usepackage{color}
\usepackage{xcolor}
\usepackage[normalem]{ulem}
\usepackage{textpos}

\title{\LARGE \bf Closing the Loop: Graph Networks to Unify Semantic Objects and  Visual Features for Multi-object Scenes}

\author{Jonathan J.Y. Kim$^{1,2*}$, Martin Urschler$^{1}$, Patricia J. Riddle$^{1}$, and J\"org S. Wicker$^{1}$ \\ $^{1}$School of Computer Science, University of Auckland, New Zealand  \:\:$^{2}$Callaghan Innovation, New Zealand
\\ {\tt\small jkim072@aucklanduni.ac.nz}, {\tt\small \{martin.urschler,p.riddle,j.wicker\}@auckland.ac.nz}
\thanks{*\textit{This research was supported by Callaghan Innovation, New Zealand's Innovation Agency}}
}

\newcommand{\etal}{{\em et al.}}

\begin{document}

\maketitle
\thispagestyle{empty}
\pagestyle{empty}

\begin{textblock*}{180mm}(.01\textwidth,-6.8cm)
Accepted at IROS 2022.
© 2022 IEEE.  Personal use of this material is permitted.  Permission from IEEE must be obtained for all other uses, in any current or future media, including reprinting/republishing this material for advertising or promotional purposes, creating new collective works, for resale or redistribution to servers or lists, or reuse of any copyrighted component of this work in other works.
\end{textblock*}


\begin{abstract}
In Simultaneous Localization and Mapping (SLAM), Loop Closure Detection (LCD) is essential to minimize drift when recognizing previously visited places. 
Visual Bag-of-Words (vBoW) has been an LCD algorithm of choice for many state-of-the-art SLAM systems. It uses a set of visual features to provide robust place recognition but fails to perceive the semantics or spatial relationship between feature points. 
Previous work has mainly focused on addressing these issues by combining vBoW with semantic and spatial information from objects in the scene. However, they are unable to exploit spatial information of local visual features and lack a structure that unifies semantic objects and visual features, therefore limiting the symbiosis between the two components. 
This paper proposes SymbioLCD2, which creates a unified graph structure to integrate semantic objects and visual features symbiotically. Our novel graph-based LCD system utilizes the unified graph structure by applying a Weisfeiler-Lehman graph kernel with temporal constraints to robustly predict loop closure candidates.
Evaluation of the proposed system shows that having a unified graph structure incorporating semantic objects and visual features improves LCD prediction accuracy, illustrating that the proposed graph structure provides a strong symbiosis between these two complementary components.
It also outperforms other Machine Learning algorithms - such as SVM, Decision Tree, Random Forest, Neural Network and GNN based Graph Matching Networks.
Furthermore, it has shown good performance in detecting loop closure candidates earlier than state-of-the-art SLAM systems, demonstrating that extended semantic and spatial awareness from the unified graph structure significantly impacts LCD performance.

\end{abstract}


\section{Introduction}
\label{sec:introduction}

Over the past three decades, Simultaneous Localization and Mapping (SLAM) has been actively adopted by wide-ranging fields, such as robotics, medical imaging and mobile devices, for simultaneously building a 3D map and finding the pose of a device. With the advent of low-cost cameras capable of obtaining clear images with relatively high framerates, there has been considerable research in visual SLAM, where SLAM is performed using only visual sensors.
Two major categories of visual SLAM are indirect SLAM \cite{ORBSLAM} and direct SLAM \cite{elasticfusion, LSDSLAM}.
Indirect or feature-based SLAM convert an image into a sparse set of visual features, such as SIFT \cite{SIFT}, SURF \cite{SURF},  or ORB \cite{ORB}, whereas direct SLAM utilizes information from all pixels, such as color and intensities \cite{feature_based, dense_map}, or edges \cite{reslam}.
 
\begin{figure}[t]
\begin{center}
\includegraphics[width=1.0\linewidth]{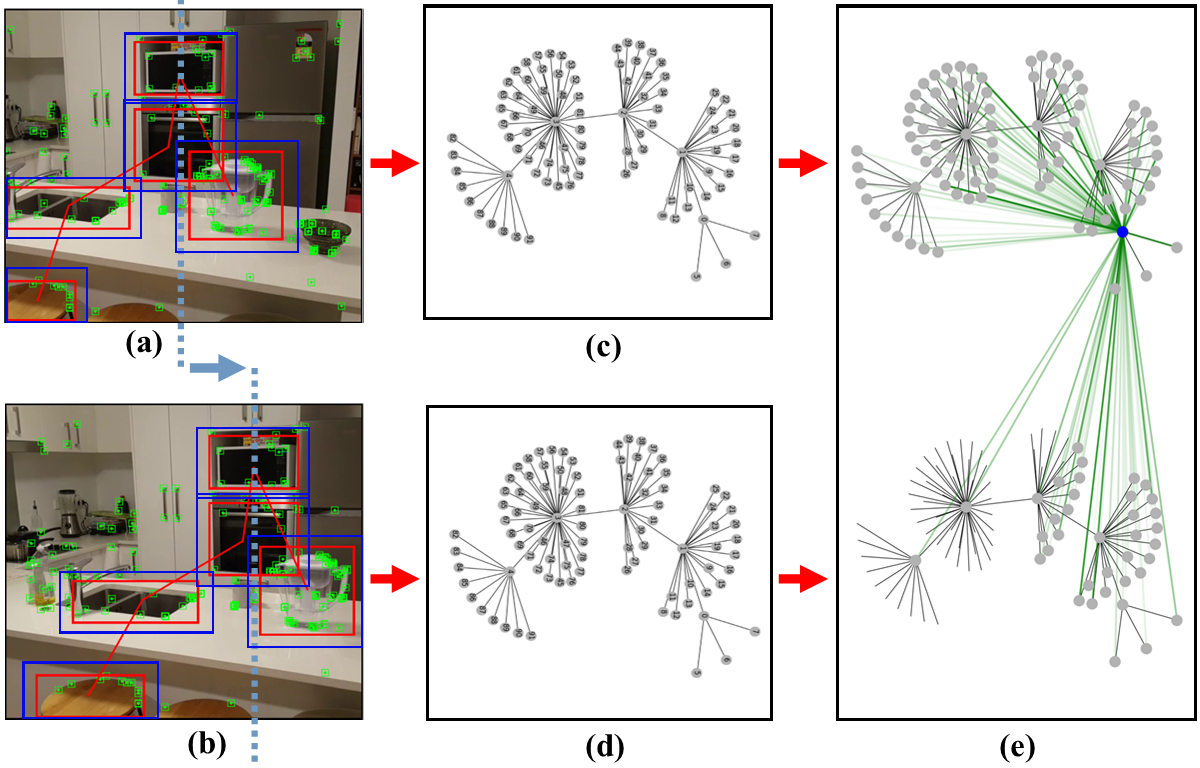}
\end{center}
\caption{Basic overview of SymbioLCD2 (a) image taken at time $i$ (b) image taken at time $j$ (c)\&(d) semantic objects and visual features are unified into a graph structure (e) GNN-based graph kernel with the subgraph matching algorithm.}
\label{fig:title}
\end{figure}

In indirect SLAM, localization and mapping rely on visual feature matching between each frame. Minor errors during feature matching and additional sensor drift from the device can accumulate over time to create a substantial amount of drift near the end of its trajectory. Monocular SLAM, where SLAM is performed using a single camera, suffers relatively larger drift compared to stereo SLAM, as it lacks a second camera to perform image rectification. This paper focuses on Monocular SLAM, as it is more crucial for monocular SLAM to address the drift.
Loop Closure Detection (LCD), also known as visual place recognition, is utilized by visual SLAM systems to minimize drift in its trajectory. 
Loop closure requires a device to return to a previously visited location and recognize features obtained earlier to determine and minimize drift accurately. 
However, LCD is a challenging task due to a phenomenon known as perceptual aliasing, where the features from two different locations may appear similar, making it difficult to ascertain whether the location has been visited previously or not.

Many SLAM systems \cite{ORBSLAM, dynaslam} utilize a visual Bag-of-Words (vBoW) algorithm \cite{BagOfWords} to perform loop closure. A common vBoW algorithm uses an offline dictionary of visual words created by encoding and clustering the features derived from a large number of images. It finds similarity in a scene by comparing the current set of visual features against feature sets found in its dictionary. 
The vBoW algorithm is fast and robust to partial occlusions, but most spatial information between feature points is lost during encoding and clustering. 

Previous work by Kim \etal\ \cite{symbio} and Zhang \etal\  \cite{Temporal_similarity} have focused on mitigating the drawback of vBoW by combining or replacing it with the semantic and spatial information from semantic objects extracted by Convolutional Neural Network (CNN). However, it was only utilizing spatial relationship between semantic objects, failing to exploit additional, potentially complementary spatial information of local visual features. It also lacked a structure that could unify both semantic objects and visual features, restricting the symbiosis between the two components.  

We propose to extend the semantic and spatial awareness of a scene by connecting semantic objects and visual features into a unified graph structure. Graph structures have become a popular medium for learning algorithms, as their non-linear data structure can represent both data and their relationships. By connecting semantic objects and their surrounding visual features, it is possible to utilize spatial information of visual features and semantic objects simultaneously, thus extending the number of spatial relationships in a scene.
The resulting graph structure can then be used to learn and predict loop closure candidates using a Graph Neural Network (GNN) \cite{GMN}. We propose to use a GNN with a dedicated graph kernel, which can predict the similarity of a pair of graphs by performing attributed subgraph matching \cite{weis}. 
The advantages of our proposed unified graph structure are as follows:
\begin{itemize}

\item Combining both semantic object and visual features into a single uniform framework, improving the symbiosis between the two components during both the learning and prediction phases of loop closure candidates. 

\item Spatial relationship in a scene can be extended by incorporating spatial information of visual features. By connecting semantic objects and their surrounding visual features, we vastly increase the number of spatial relationships that can be extracted from the scene. 

\item Semantic information of the object can be extended by sharing the semantic label with its surrounding visual features. A semantic object, such as a cup, is no longer just a label, as it also incorporates distinct relational connections from its surrounding visual features, thereby extending the semantic information of the scene.

\end{itemize}

This paper presents SymbioLCD2, a novel graph-based loop closure detection system that uses a unified graph structure and a GNN with graph kernel for a robust loop closure candidate detection. It combines the strengths of both semantic objects and visual features symbiotically by transforming them into a unified multi-tier graph structure. Figure 1 shows the basic framework, and Figure 2 shows a detailed overview of our proposed SymbioLCD2.

The main contributions of this paper are as follows:
\begin{enumerate}

\item A novel unified graph structure combining semantic objects and visual features symbiotically, providing improved semantic and spatial awareness of a scene.

\item Multi-tiered graph formation algorithm with object anchors.

\item A novel Loop Closure Detection system using Graph Kernel with temporal constraints for a robust and accurate loop closure candidate prediction.

\item Better precision and recall rate compared to other ML algorithms.

\item Earlier detection of loop closure candidates than state-of-the-art SLAM system.

\end{enumerate}

This paper is organized as follows: Section \ref{sec:relatedwork} reviews related work, Section \ref{sec:proposedmethod} describes our proposed method, Section \ref{sec:experiments} presents experiments and their results, and Section \ref{sec:conclusion} presents the conclusion and future work.
\begin{figure*}[ht]
\begin{center}
\includegraphics[width=1.0\linewidth]{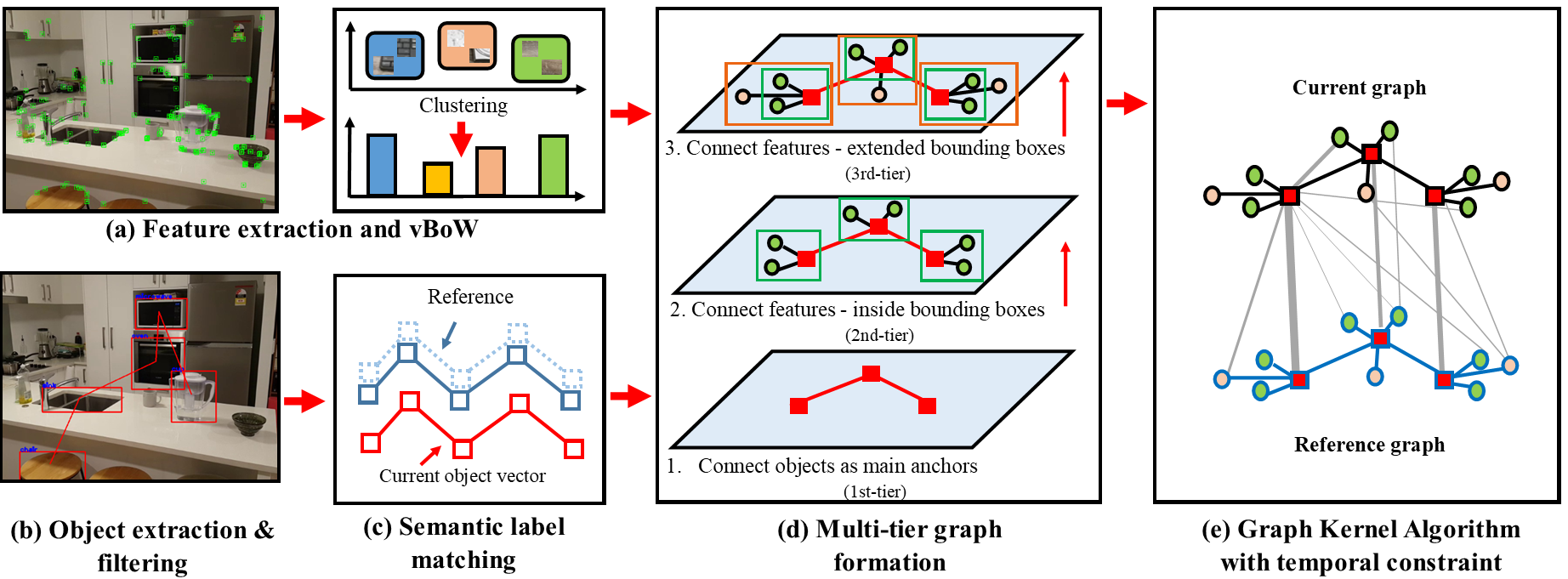}
\end{center}
   \caption{Detailed overview of SymbioLCD2. (a) visual feature extraction and generating a vBoW score (b) semantic object extraction from CNN and object filtering based on their location and size (c) normalized semantic label matching (d) multi-tier graph formation using semantic objects as main anchors - object and feature information gets transferred as node features and distances between objects and features become edge features (e) graph kernel with the subgraph matching algorithm - temporal constraints are added to penalize frames that are close to each other. }
\label{fig:det_overview}
\end{figure*}
\section{Related Work}
\label{sec:relatedwork}
This section reviews feature descriptors, visual Bag-of-Words, CNN derived descriptors, and Graph Neural Network algorithms.

\textbf{Feature descriptors}
Feature descriptors \cite{Feature_matching}, such as SURF \cite{SURF} and SIFT \cite{SIFT}, are algorithms used to extract the visual features from input images. They have been widely used in feature-based SLAM systems, but more recently binary descriptors, such as ORB \cite{ORB} and BRIEF \cite{BRIEF}, have been used in various state-of-the-art SLAM systems due to their computational advantage. In general, traditional feature descriptors offer better accuracy, but binary descriptors are faster and more efficient \cite{Brisk_comp}. 

\textbf{Visual Bag-of-Words} 
Visual Bag-of-Words are commonly found in loop closure detection modules in several state-of-the-art SLAM systems, such as ORB-SLAM \cite{ORBSLAM}, LSD-SLAM \cite{LSDSLAM} and DynaSLAM \cite{dynaslam}. 
There are two distinct vBoW approaches. FAB-MAP \cite{FABMAP} learns a generative appearance model using Chow-Liu tree, and it can perform fast online loop closure detection due to the algorithm's linear complexity. DBoW2 \cite{DBOW2}, on the other hand, creates a tree-structured offline dictionary by training on a large set of image datasets. DBoW2 compares a set of features extracted by feature descriptors against its dictionary to calculate the co-occurrence score.

\textbf{CNN derived descriptors for SLAM}
With the recent advancement in CNNs, there has been a significant interest in using CNNs to extract semantic objects from images and videos.
Object instance segmentation algorithms such as Faster R-CNN \cite{RCNN}, Mask R-CNN \cite{maskrcnn}, and YOLO \cite{YOLO3} can extract semantic and spatial information of objects from a scene and generate semantic labels, bounding boxes and masks of objects. The CNN-extracted objects have been successfully used to replace visual features \cite{seman_map} in such SLAM systems as NodeSLAM \cite{nodeslam} and Quadric SLAM \cite{Quadslam}. 

\textbf{Graph Neural Networks with Graph Kernel}
Graph structured data can be an effective format for encoding relational spatial structures in a scene. In recent years, GNN have been widely adopted for efficiently learning relational representations in graph-structured data \cite{graphsage}. GNN models are commonly used for graph classification \cite{ete}, where they can predict the similarity of a graph pair by producing graph embeddings in vector spaces for efficient similarity reasoning \cite{GMN, GAN}. Recent work on graph kernels has emerged as a promising approach for graph classification, as they allow kernelized learning algorithms, such as SVM and Weisfeiler-Lehman, to perform attributed subgraph matching, achieving state-of-the-art performance on graph classification tasks \cite{grakel}.  Weisfeiler-Lehman (WL) graph kernel uses a distinctive labelling scheme to extract subgraph patterns through multiple iterations. It replaces the label of each node with a label consisting of its original label and the subset of labels of its direct neighbours, then compresses it to form a new label. After the relabeling procedures, it calculates the similarity of two graphs as the inner product of the histogram vectors of both graphs \cite{weis}.

We combine semantic objects extracted from Mask R-CNN and visual features extracted from ORB descriptor to form a unified graph structure to enhance the symbiosis between two components, and perform graph classification using the WL graph kernel.

\section{Proposed Method}
\label{sec:proposedmethod}
The first three parts in our methods - \ref{sec:framework}, \ref{sec:objextraction} and \ref{sec:labelmatching}, are identical to the work presented in \cite{symbio}.
Our proposed methods are as follows - first, visual features are extracted by ORB feature descriptor and the vBoW score is generated using DBoW2.
Second, semantic objects are extracted using Mask R-CNN then filtered based on their semantic labels and size. 
Third, the filtered objects get transferred into a matrix, then the semantic and spatial information in the matrix gets projected onto a normalized plane for a scale-invariant label distance matching. 
Fourth, objects in the matrix get connected to form the main anchors of the graph. Any visual features within the bounding box of an anchor object are assigned with the semantic label of the object and then connected to their anchor object to form a two-tier graph. Then we extend the graph to include surrounding visual features within $x$ pixels from the edges of the bounding box to complete a three-tier graph.
Lastly, the combined graph is transferred to a WL graph kernel with temporal constraints for loop closure candidate prediction.
The overview of our proposed method is shown in Figure 2, and the reader may refer to \cite{symbio} for further details on (a), (b) and (c).
\subsection{Framework, visual feature extraction and generating a vBoW score}
\label{sec:framework}
SymbioLCD2 is built upon SymbioSLAM2 as its operating framework, which is based on DynaSLAM \cite{dynaslam}, and it incorporates ORB feature descriptor, DBoW2 and Mask R-CNN within its framework. We use ORB feature descriptor to extract visual features and DBoW2 to compute the vBoW score.
\subsection{Object extraction and filtering process}
\label{sec:objextraction}
Semantic objects are extracted using Mask R-CNN, then filtered based on their semantic labels. Any objects with labels that correspond to movable entities, e.g. person, car or bicycles, are filtered out. This filtering process reduces an unintended loop closing on objects that are not stationary, thus improving the robustness of loop closure.
Additionally, any objects that are too large and overlap with other objects in the scene are also filtered out. 
\subsection{Semantic label matching in a scale invariant normalized plane}
\label{sec:labelmatching}
Aligning and matching object labels based on their raw distances from each other can cause errors due to changes in scale and viewpoint between frames. By projecting object representations into a normalized plane, the distances between a pair of objects become scale-invariant, assisting the label matching algorithm to identify their match accurately.
It also enables the label matching algorithm to allow up to 40\% of missing or misclassified labels, making it cope better with detection errors originating from the previous object extraction process.

\subsection{Graph structure formation - three-tier graph formation with central object anchor nodes}
\label{sec:graph_str}
\begin{figure}[t]
\begin{center}
\includegraphics[width=1.0\linewidth]{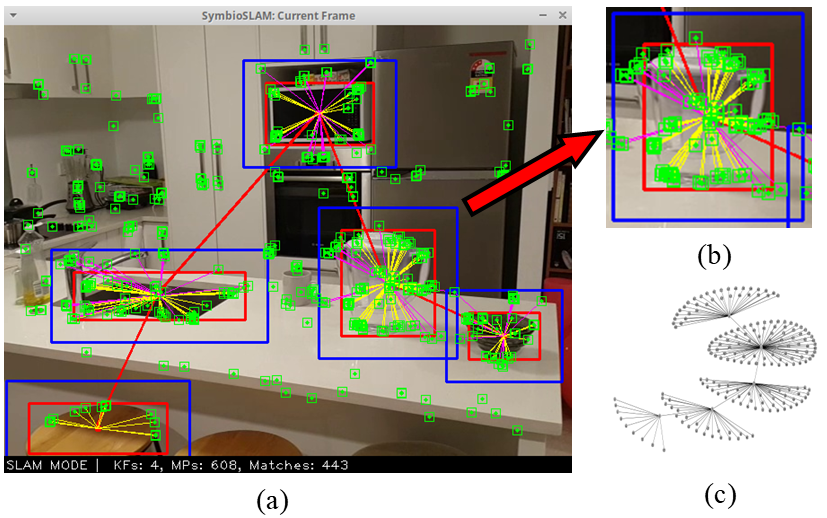}
\end{center}
\caption{Three-tier graph connection (a) graph representation in SLAM (b) first-tier connections are shown in red, second-tier connections are in yellow and third-tier connections are in magenta (c) graph representation in NetworkX with normalized distances.}
\label{fig:vbow_conn}
\end{figure}

We have designed our graph structure to have three different tiers. 
The three-tier formation of the graph provides a stable structure for the graph, which helps to improve the accuracy of loop closure candidate prediction.
Figure \ref{fig:det_overview}(d) shows an overview of our graph structure formation, and Figure \ref{fig:vbow_conn} shows an example of a three-tier graph represented in SLAM and in NetworkX \cite{networkx}.

The first-tier connects objects to form central nodes that act as anchors for visual features to connect onto in the upper tiers. The anchor allows the graph to extend the semantic information of the object by propagating it to the overall graph structure. The object information and the vBoW score get assigned as node features. The normalized distance between each pair of objects is assigned as the edge feature.

The second-tier connects visual features within bounding boxes to each anchor node, i.e. the bounding box's object node. This multi-tier connection provides extra symbiosis in the graph structure. First, the semantic object gets encoded with positional information from its surrounding visual features, extending its spatial information. Second, visual features get assigned with the label from the semantic object, thereby extending the semantic awareness of the scene - the visual features can now be associated with semantic objects and obtain semantic information through their connections, allowing them to be more than just feature points in the background. The vBoW score and the semantic label of its anchor node get assigned as node features. The normalized distance between the visual feature to its anchor node is assigned as the edge feature.

The third-tier extends the graph to include surrounding visual features within $x$ pixels from the edges of the bounding box. For the experiments, $x$ was set to 25 pixels to minimize the interference with neighbouring objects. The third-tier allows the graph to include the visual features from the object's background, further enhancing the spatial awareness of the scene. The features in the third-tier do not get assigned with the semantic labels since they are not part of the semantic object, but it helps the graph kernel to differentiate semantic objects when there are multiple similar objects present in the scene.

\begin{figure*}[ht]
\begin{center}
\includegraphics[width=0.9\linewidth]{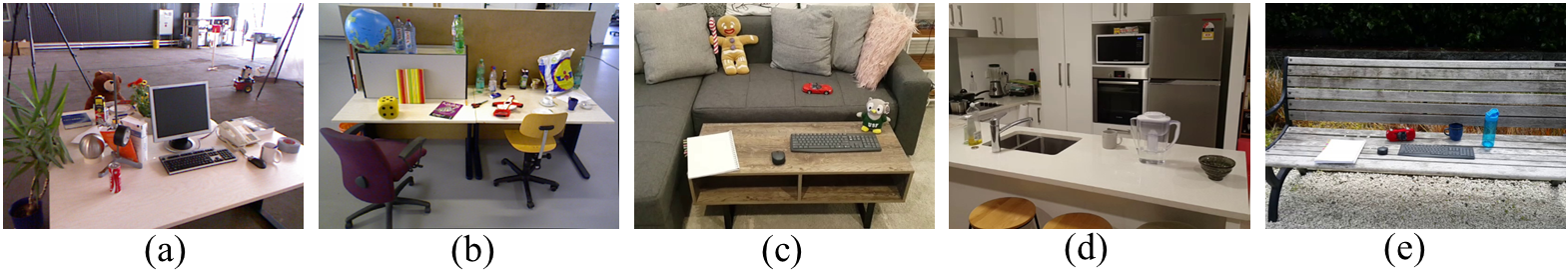}
\end{center}
  \caption{Evaluation datasets. (a) fr2-desk (b) fr3-longoffice (c) uoa-lounge (d) uoa-kitchen (e) uoa-garden}
  \label{fig:datasets}
\end{figure*}

\subsection{LCD prediction using GNN - Weisfeiler-Lehman subgraph kernel with temporal constraints}
\label{sec:lcdprediction}
The three-tier graph structure, with combined information from semantic objects and visual features, gets transferred to the graph kernel LCD predictor. 
We use Weisfeiler-Lehman (WL) subgraph algorithm \cite{weis} as our GNN algorithm, with a vertex histogram as its base kernel, and a multi-layer perceptron at the end for the loop closure candidate detection. 

The WL subgraph kernel takes a pair of graphs, i.e. a graph from the current frame and a graph from the reference frame, and computes their similarity. Unlike typical GNN embedding models \cite{graphsage} \cite{GAN}, the WL subgraph matching computes the similarity score jointly on the pair, rather than first mapping each graph to a set of graph embeddings. The key idea of the WL subgraph matching algorithm is to substitute the label of each node with a label that includes its original label and the subset of labels from its neighbors, then compressing it to form a new multi-set label. This procedure gets repeated for \textit{h} iterations and is performed on the pair of graphs simultaneously, therefore any two groups of vertices, or subgraphs, gets identical new labels only if they have identical multi-set labels.

Given two graphs $G^i$ and $G^j$, with \textit{i} \& \textit{j} as the indices of two images being compared,

\begin{equation}
k_{\text{WL}}(G^i,G^j) = k(G^i_0, G^j_0)+ k(G^i_1, G^j_1)+...+k(G^i_h, G^j_h),
\end{equation}

where $h$ is the number of iterations and $k$ being the base kernel for WL, in our case the vertex histogram kernel. 

For finding similarity using subgraph information, let $\sum_0$ be the set of original node labels of $G^i$ and $G^j$, and $c_h(G^{ij},\sigma_{ij})$ as the number of occurrences of the label $\sigma_{ij}$ in the Graph $G^i$ and $G^j$ after $h$ iterations.

Then the Weisfeiler-Lehman subgraph kernel can be defined as:
\begin{equation}
k_{\text{WL}sg}(G^i,G^j) = (\theta^{h}_{\text{WL}sg}(G^i) , \theta^{h}_{\text{WL}sg}(G^j)),
\end{equation}

where

\begin{equation}
\theta^{h}_{\text{WL}sg}(G^i)=(c_0(G^{i},\sigma_{0|\sum_0|}),...,c_h(G^{i},\sigma_{h|\sum_h|)}),
\end{equation}

and 

\begin{equation}
\theta^{h}_{\text{WL}sg}(G^j)=(c_0(G^{j},\sigma_{0|\sum_0|}),...,c_h(G^{j},\sigma_{h|\sum_h|)}).
\end{equation}

Since SLAM uses a continuous stream of images as its input, adjacent images in the sequence can look very similar to each other. Therefore, we use a modified temporal similarity constraint first proposed by Zhang \etal\ \cite{Temporal_similarity}, 

\begin{equation}
Tc(i,j) = \ln\left (\frac{\beta_sv^2_c}{f_c}(i-j)^2\right ),
\end{equation}

where \textit{v\textsubscript{c}} is the velocity of the camera, $\beta$\textsubscript{s} is a constant parameter and \textit{f\textsubscript{c}} is the frame rate.
To simplify, we set \textit{v\textsubscript{c}} and \textit{f\textsubscript{c}} to be constant, giving us

\begin{equation}
Tc(i,j) = \ln(\beta_s(i-j)^2),
\end{equation}

where
\begin{equation}
\beta_s \: \epsilon \: (0,1).
\end{equation}

We combine the temporal constraints with WL subgraph kernel to create a similarity equation $S(i,j)$, written as

\begin{equation}
S(i,j) =k_{\text{WL}sg}(G^i,G^j) - \alpha(Tc(i,j)),
\end{equation}

where $\alpha  \: \epsilon \: (0,100) $ is a parameter to control the weight of $Tc(i,j)$.
When the frame i and j are close the $Tc(i,j)$ will be significant, thus decreasing the WL similarity score. As the $i$ and $j$ get further apart, the value of the constraint will continue to decrease until it becomes negligible, thus not affecting the similarity score.

\section{Experiments}
\label{sec:experiments}
We have evaluated SymbioLCD2 with the following experiments. 
Section \ref{sec:datasets} shows evaluation parameters and datasets used in the experiments.
Section \ref{sec:measuringsymbiosis} performs ablation studies, evaluating symbiosis between semantic objects and visual features in our proposed three-tier graph.
Section \ref{sec:precision} evaluates our graph-based LCD predictor against other ML algorithms. Section \ref{sec:keyframe} compares the performance of LCD candidate detection against state-of-the-art SLAM systems.


\subsection{Datasets}
\label{sec:datasets}
For evaluating our SymbioLCD2, we have selected five publicly available datasets with multiple objects and varying camera trajectories. 
We selected fr2-desk and fr3-longoffice from the TUM dataset \cite{TUM_data}, and uoa-lounge, uoa-kitchen and uoa-garden from the University of Auckland multi-objects dataset \cite{symbio}. Table \ref{tab:eval} shows evaluation parameters and Figure \ref{fig:datasets} shows examples from each dataset. All experiments were performed on a PC with Intel i9 7900X and Nvidia GTX1080Ti.

\begin{table} [b]
\centering
\caption{The evaluation parameters}
\begin{tabular}{ccccccc} \toprule \midrule
      $x$ & $\beta_s$ & $\alpha$  & $h$ & edge dim. & node dim. & base kernel \\ \midrule
      25 & 0.3  & 2 & 50 & 4  & 8  & Vertex Histogram \\\midrule \bottomrule
\label{tab:eval}
\end{tabular}
\end{table}


\begin{table*} [ht]
\centering
\caption{Evaluation against other ML algorithms}
\label{tab:precision}
\begin{tabular}{ccccccccccccc} \toprule \midrule
      
    &\multicolumn{2}{c}{SVM-RBF}
    &\multicolumn{2}{c}{DecisionTree}
    &\multicolumn{2}{c}{NeuralNetwork}
    &\multicolumn{2}{c}{SymbioLCD}
    &\multicolumn{2}{c}{GraphMatchingNet}
    &\multicolumn{2}{c}{SymbioLCD2}
   \\
      Dataset &Precision&Recall &Precision&Recall &Precision&Recall &Precision&Recall &Precision&Recall &Precision&Recall
      
      \\ \midrule
      fr2-desk       &90.91&83.33 &\textbf{100.00}&75.38 &76.92&83.33 &92.31&\textbf{100.00} &91.91&92.36 &96.77&97.27
      \\
      fr3-longoffice  &67.74&77.78 &95.24&76.92 &66.67&88.89 &96.30&94.43 &92.86&90.66 &\textbf{97.22}&\textbf{97.81}
      \\
      uoa-lounge    &85.00&83.95 &93.55&77.33 &86.67&96.30 &98.77&97.57 &94.88&93.76 &\textbf{99.35}&\textbf{98.34}   
      \\
      uoa-kitchen   &89.73&83.44 &93.16&77.53 &91.07&97.45 &\textbf{99.32}&93.63 &94.05&96.42 &98.92&\textbf{97.73}
      \\
      uoa-garden     &89.04&86.28&94.90&79.01&90.20&97.79&\textbf{99.54}&95.13&95.08&96.77&99.16&\textbf{98.99}
      \\ \midrule 
      Average    &84.48&82.96&95.37&77.24&82.31&92.75&97.25&96.16&93.76 &93.99&\textbf{98.28}&\textbf{98.03}
      \\
      \bottomrule

\end{tabular}
\end{table*}

\subsection{Ablation studies - evaluating symbiosis between semantic objects and visual features}
\label{sec:measuringsymbiosis}
\begin{figure}[t]
\begin{center}
\includegraphics[width=1.0\linewidth]{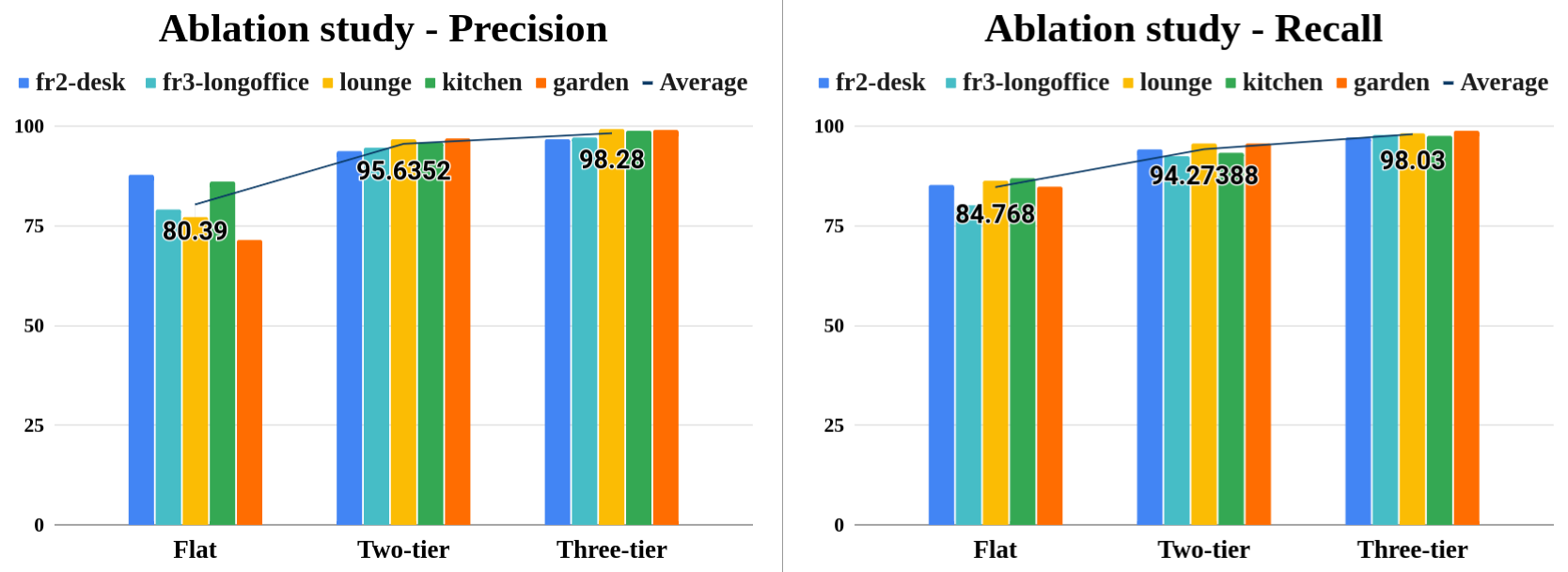}
\end{center}
   \caption{Ablation study - Precision and Recall graphs for Flat (visual features only) vs. Two-tier vs. Three-tier}
   \label{fig:ablation}
\end{figure}

We have performed ablation studies to analyze two aspects of our proposed method. First, measuring the symbiosis between semantic objects and visual features by comparing it against a graph composed with only visual features.
Second, determining whether extended spatial information from the third-tier of the graph, i.e. the surrounding visual features from the background, improves the prediction accuracy. 
For this experiment, we created a flat structured graph with just visual features and a two-tier graph with no visual features from the background, to compare against our three-tier graph structure. 
The evaluation was measured using precision and recall metric, which are defined as follows,
\begin{equation}
Precision= \frac{TP}{TP+FP}  \;\;\; \& \;\;\; Recall= \frac{TP}{TP+FN},
\end{equation}
where TP refers to true positive, FP refers to false positive and FN for false negative.
Figure \ref{fig:ablation} and Table \ref{tab:ablation} show that two-tier graphs with semantic objects outperform the flat graph by 15.44\% improvements in precision and 9.5\% improvements in recall. This demonstrates that having semantic objects as part of the structure brings positive improvements to the prediction performance.
The result also shows that having a three-tier structure, which includes surrounding visual features from the background, improves the prediction performance by 2.45\% in precision and 3.77\% in recall compared to a two-tier graph. This demonstrates that extended spatial awareness from visual features in the third-tier contributes to improvements in performance.
The results show that through the use of the proposed framework, the strong symbiotic relationship can be effectively utilized, where both semantic objects and visual features contribute to higher LCD prediction accuracy.

\subsection{Evaluating LCD keyframe prediction against other ML algorithms}
\label{sec:precision}
We have benchmarked SymbioLCD2 against five other Machine Learning algorithms - GNN based Graph Matching Network (GMN), SVM-RBF, DecisionTree, Neural Network with 4x16 layers and SymbioLCD with Random Forest classifier. We used a multi-tier graph structure to evaluate SymbioLCD2 and GMN, but used tabular data, i.e. object labels, normalized label matching, Hausdorff distance and vBoW scores, for other algorithms as they could not take a graph structure as their input. In this evaluation, we measured the performance of each algorithm by predicting keyframes of loop closure candidates and used precision and recall metrics for the evaluation.

\begin{table} [t]
\centering
\caption{Ablation study - Flat vs. Two-tier vs. Three-tier}
\label{tab:ablation}
\begin{tabular}{ccccccc} \toprule \midrule
      
    &\multicolumn{2}{c}{Flat}
    &\multicolumn{2}{c}{Two-tier}
    &\multicolumn{2}{c}{Three-tier}

   \\
      Dataset &Preci.&Recall &Preci.&Recall &Preci.&Recall
      
      \\ \midrule
      fr2-desk       &87.92 &85.33 &93.74 & 94.20 &\textbf{96.77} &\textbf{97.27} 
      \\
      fr3-longoffice  &79.12 &80.28 &94.71 &92.47 &\textbf{97.22} &\textbf{97.81} 
      \\
      uoa-lounge        &77.23 &86.32 &96.77 &95.63 &\textbf{99.34} &\textbf{98.34} 
      \\
      uoa-kitchen       &86.23 &87.08 &95.93 &93.34 &\textbf{98.92} &\textbf{97.73} 
      \\
      uoa-garden        &71.45 &84.83 &96.98 &95.70 &\textbf{99.16} &\textbf{98.99} 
      \\ \midrule 
      Average       &80.39 &84.76 &95.83 &94.26 &\textbf{98.28} &\textbf{98.03} 
      \\
      \bottomrule

\end{tabular}
\end{table}

\begin{table}[t]
\centering
\caption{Autorank analysis (Precision)}
\label{tab:aut_pre}
\begin{tabular}{lrrllll}
\toprule
{} &    MR &    MED &   MAD &    CI & $\gamma$ &   Mag. \\
\midrule
\textbf{SymLCD2}    & 1.50 & 98.60 & 0.97 &  [98.28, 98.92] &     0.0 &  neg. \\
SymLCD     & 1.83 & 98.01 & 2.11 &  [97.25, 98.77] &     0.3 &       small \\
DecTree       & 3.16 & 95.07 & 1.34 &  [94.90, 95.24] &     3.0 &       large \\
GMN     & 3.50 & 93.90 & 1.49 &  [93.76, 94.05] &     3.7 &       large \\
SVM-RBF       & 5.50 & 87.02 & 3.89 &  [85.00, 89.04] &     4.0 &       large \\
NeuralNet    & 5.50 & 84.49 & 9.11 &  [82.31, 86.67] &     2.1 &       large \\
\bottomrule
\end{tabular}
\end{table}

\begin{table}[t]
\centering
\caption{Autorank analysis (Recall)}
\label{tab:aut_rec}
\begin{tabular}{lrrllll}
\toprule
{} &    MR &    MED &   MAD &        CI & $\gamma$ &   Mag. \\
\midrule
\textbf{SymLCD2}    & 1.16 & 97.92 & 0.45 &  [97.81, 98.03] &     0.0 &  neg. \\
SymLCD     & 2.50 & 95.64 & 2.32 &  [95.13, 96.16] &     1.3 &       large \\
GMN         & 3.16 & 93.87 & 3.01 &  [93.76, 93.99] &     1.8 &       large \\
NeuralNet    & 3.25 & 94.52 & 4.58 &  [92.75, 96.30] &     1.0 &       large \\
SVM-RBF       & 4.91 & 83.38 & 0.73 &  [83.33, 83.44] &    2.8 &       large \\
DecTree       & 6.00 & 77.28 & 0.45 &  [77.24, 77.33] &    4.6 &       large \\

\bottomrule
\end{tabular}
\end{table}
\begin{figure}[b]
\begin{center}
\includegraphics[width=1.0\linewidth]{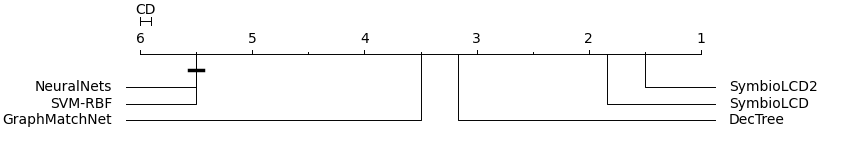}
\end{center}
  \caption{Critical Difference diagram (Precision) }
  \label{fig:autorank_pres}
\end{figure}

\begin{figure}[b]
\begin{center}
\includegraphics[width=1.0\linewidth]{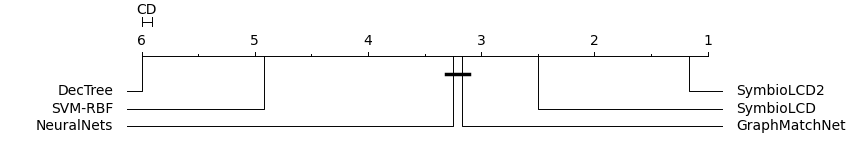}
\end{center}
  \caption{Critical Difference diagram (Recall) }
  \label{fig:autorank_rec}
\end{figure}

  
The evaluation was performed 50 times to account for the non-deterministic nature of the algorithms. 
Table \ref{tab:precision} presents the precision and recall values of SymbioLCD2 against other algorithms in each dataset. The result shows that, on average, SymbioLCD2 achieved the highest precision and recall compared to other algorithms. 

To examine the statistical robustness of results, we have used Autorank \cite{autorank} to analyze each algorithm's performance further. Autorank is an automated classifier ranking algorithm that is based on the guidelines proposed by Demšar \cite{cd_diagram}.
It uses paired samples that are independent of each other and determines the differences in the central tendency, such as median (MED), mean rank (MR) and median absolute deviation (MAD), to rank each algorithm.
Table \ref{tab:aut_pre} and Figure \ref{fig:autorank_pres} shows that SymbioLCD2 was ranked highest against other ML algorithms in precision, Table \ref{tab:aut_rec} and Figure \ref{fig:autorank_rec} shows that SymbioLCD2 ranked highest in recall.

\subsection{Evaluating LCD keyframe prediction against other SLAM systems}
\label{sec:keyframe}

To evaluate the performance of SymbioLCD2, we have benchmarked it against the state-of-the-art ORB-SLAM2, DynaSLAM and SymbioLCD. We have chosen the above mentioned three systems for benchmarking, as they all share the same vBoW algorithm and keyframe insertion algorithm. 
In this evaluation, we recorded the earliest keyframe number of loop closure candidates to benchmark their performances. We have performed the evaluation five times and the median value was recorded to account for the non-deterministic nature of the system.
The result in Table \ref{tab:keyframe} shows that SymbioLCD2 outperformed ORB-SLAM2, DynaSLAM and SymbioLCD in all datasets. On average, SymbioLCD2 outperformed ORB-SLAM2 by 18.2 frames (3.98\%), DynaSLAM by 22.4 frames (6.18\%) and SymbioLCD by 3.2 frames (1.10\%). 
This evaluation demonstrates that having extended semantic and spatial awareness contributes to the early detection of loop closure candidates.

\begin{table} [h]
\centering
\caption{Comparisons of loop closure detected keyframe.}
\label{tab:keyframe}
\begin{tabular}{ccccc} \toprule \midrule
      Dataset & \vtop{\hbox{\strut ORB}\hbox{\strut SLAM2}\hbox{\strut (kf)}} &
      \vtop{\hbox{\strut Dyna}\hbox{\strut SLAM}\hbox{\strut (kf)}} & 
      \vtop{\hbox{\strut Symbio}\hbox{\strut LCD}\hbox{\strut (kf)}} &
      \vtop{\hbox{\strut Symbio}\hbox{\strut LCD2}\hbox{\strut (kf)}}
      
      \\ \midrule
     fr2-desk           & 393 & 397 & 388           & \textbf{386}\\
     fr3-longoffice     & 345 & 349 & 314  & \textbf{312} \\
     uoa-lounge             & 301 & 303 & 284           & \textbf{279}\\
     uoa-kitchen            & 410 & 416 & 392           & \textbf{385} \\
     uoa-garden             & 454 & 459 & 450           & \textbf{446} \\ 
\midrule \bottomrule

\end{tabular}
\end{table}

\section{Conclusion and Future Work}
\label{sec:conclusion}
We presented SymbioLCD2, a graph-based loop closure detection system using a graph kernel with temporal constraints for a robust and accurate loop closure candidate prediction. We presented a three-tier graph structure with object anchors to extend semantic and spatial awareness of the scene, connecting semantic objects and visual features.
We showed that our unified graph structure helps in making effective use of the strong symbiosis between semantic objects and visual features, where both components contribute to improving the performance of loop closure candidate prediction. SymbioLCD2 outperformed other ML algorithms in both precision and recall, and detected loop closure candidates earlier than state-of-the-art SLAM systems. SymbioLCD2 requires multiple static objects in a scene to be most efficient. For future research, we aim to extend our SymbioLCD2 to work with both static and dynamic objects by utilizing 3D components from a stereo or depth camera.

{\small
\bibliographystyle{IEEEtran}
\bibliography{egbib}
}

\end{document}